\documentclass[9pt]{article}
\usepackage{spconf,amsmath,amssymb,graphicx,hyperref}
\usepackage{booktabs}
\usepackage{makecell}
\usepackage{multirow}
\usepackage{threeparttable}
\usepackage{tabularx}
\usepackage{xcolor}
\usepackage{subcaption}

\title{Sequential Adapter Stacking for Cross-Lingual Low-Resource ASR}
\name{Thai Thi Thanh Thao Dang, Mengjie Qian, Kate Knill}
\address{Department of Engineering, University of Cambridge, UK}
\begin{document}
%\ninept
%
\maketitle
\begin{abstract}
Extending large-scale multilingual automatic speech recognition (ASR) models to low-resource languages remains challenging. Model performance is skewed toward high-resource languages and degrades sharply for languages with limited labeled data and pre-training exposure. To address this, we investigate parameter-efficient approaches for transferring knowledge from resource-rich source languages to low-resource target languages on Whisper. Alongside warm initialization and attention-based fusion, we propose Sequential Adapter Stacking, which places a trainable target-language adapter on top of a frozen source-language adapter. Under controlled experiments, these approaches are evaluated on three target languages unsupported by Whisper -- Asturian, Assamese, and Xhosa -- using source languages with varying degrees of relatedness. Sequential Adapter Stacking with the closest related source consistently and significantly outperforms full fine-tuning across the three targets, with 5--8\% relative WER reductions. These gains largely persist with only one hour of target training data.
\end{abstract}
%

% ----------------------------------------
\section{Introduction}
\label{sec:intro}
% ----------------------------------------

% Motivation
Large-scale multilingual speech models such as Whisper~\cite{radford2022robust} and MMS~\cite{pratap2024mms} have markedly advanced Automatic Speech Recognition (ASR) across languages. However, performance remains highly uneven across languages and deteriorates sharply for languages absent from pre-training. Adapting such models to unsupported languages is particularly challenging when limited transcribed speech is available. Moreover, adaptation performance depends not only on the target language being seen during pre-training, but also on the model's exposure to related languages~\cite{rouditchenko2023comparison}, suggesting that representations learned from related languages could provide useful knowledge for target-language adaptation.

Full fine-tuning is a straightforward approach to language adaptation, but it updates the entire model and risks overwriting useful pre-trained representations when target data is scarce~\cite{qian2024learn}. Parameter-efficient fine-tuning such as adapter tuning \cite{houlsby2019parameter, pfeiffer2020adapterhub}, Low-Rank Adaptation (LoRA) \cite{hu2022lora}, prefix tuning \cite{liliang2021prefix}, and prompt tuning \cite{lester2021prompt} instead adapt a small amount of parameters while keeping the pre-trained backbone frozen. Bottleneck adapters are particularly attractive, since independently trained language adapters can be reused and composed for cross-lingual transfer from resource-rich \emph{source} languages to low-resource \emph{targets}~\cite{liu2024parameter,wang2025low}. 
Existing cross-lingual ASR approaches exploit source adapters in different ways. Dual-adapter methods combine language-specific and language-general modules \cite{winata2021adapt}, warm initialization transfers source knowledge by initializing target adaptation from a source-language module \cite{song2024lora}, and fusion-based methods combine information from multiple language adapters dynamically through learned weighting \cite{hou2021exploit, hu2024langfusion,hucam2024crosslingual}. Despite these developments, how best to exploit a pre-trained source-language adapter when adapting a massively multilingual ASR model to an unsupported language remains under-explored.

A related question is which source language should be used. Language selection has historically received more attention in Natural Language Processing (NLP) than speech recognition. In NLP, no single feature reliably identifies the best source across tasks, and features grounded in the target dataset (e.g. subword overlap) and the model's own representations (e.g. embedding similarity) have proven highly predictive of cross-lingual transfer \cite{lin2019choosing,eronen2023zeroshot,idris2026embedding}. In speech recognition, previous cross-lingual ASR studies commonly select source languages based on availability or linguistic intuition~\cite{song2024lora,hou2021exploit,hu2024langfusion,hucam2024crosslingual}. Even when language similarity is quantified beforehand, it is used to predict transfer outcomes under a fixed pipeline rather than to drive an adaptation decision~\cite{wu2021crosslingual}, and phonetic metrics alone are unreliable predictors~\cite{farooq2022investigating}. This motivates jointly examining both how source knowledge is transferred and how source--target relatedness affects that transfer.

In this work, Whisper serves as a case study of a massively multilingual ASR backbone. Inspired by the modular architecture of MAD-X \cite{pfeiffer2020mad}, we propose Sequential Adapter Stacking (SeqStack) for cross-lingual source-adapter transfer and use three complementary language similarity measures to rank candidate sources before adaptation. The approach is compared with existing transfer strategies, including warm initialization and weighted fusion, on Asturian, Assamese, and Xhosa, which span different levels of related-language pre-training coverage in Whisper. The contributions of this work are as follows: 
(1) the first evaluation, to our knowledge, of SeqStack for cross-lingual ASR on a massively multilingual backbone, where it yields statistically significant improvements over full fine-tuning and monolingual adapter tuning across all three target languages; (2) a controlled comparison of three source-adapter transfer strategies, i.e. warm initialization, attention-based fusion, and SeqStack, under matched training conditions; (3) a practical framework for extending multilingual ASR to unsupported languages by combining similarity-based source selection with SeqStack.

% ----------------------------------------
\section{METHODS}
\label{sec:method}
% ----------------------------------------

This section reviews the source-adapter transfer approaches compared in our evaluation (Section~\ref{ssec:adapter-composition}), presents \textbf{Seq}uential Adapter \textbf{Stack}ing (Section~\ref{ssec:seqstack}), and describes the three complementary language similarity measures used to select source languages prior to adaptation (Section~\ref{ssec:langsec}).
Throughout, $t$ denotes the target language, and $\mathcal{S} = \{s_1, \ldots, s_K\}$ is the set of auxiliary source languages. We use $\Phi_\ell$ to represent a bottleneck adapter for language $\ell$, inserted into the frozen Whisper backbone with pre-trained weights $\Theta$, and $\mathcal{D}_\ell$ is the corresponding training set.

% ----------------------------------------
% \subsection{Compared Adapter Composition Approaches}
\subsection{Compared Source-Adapter Transfer Approaches}
\label{ssec:adapter-composition}
% ----------------------------------------
% Adapter
Bottleneck adapters \cite{houlsby2019parameter} insert a down-projection $\mathbf{W}_{\mathrm{down}} \in \mathbb{R}^{d_{\mathrm{model}} \times d}$, a non-linear activation $f$, and an up-projection $\mathbf{W}_{\mathrm{up}} \in \mathbb{R}^{d \times d_{\mathrm{model}}}$ into each Transformer layer, with bottleneck dimension $d \ll d_{\mathrm{model}}$. Following the Pfeiffer configuration \cite{pfeiffer2020adapterhub}, a single adapter is applied to the Feed-Forward Network output $\mathbf{h}$ in each Whisper encoder and decoder layer:\footnote{All adapters use a zero-initialized up-projection, as the  default initialization degraded Whisper's auto-regressive generation in initial experiments.}
% \vspace{-2mm}
\[
    \mathrm{Adapter}(\mathbf{h}) = \mathbf{h} + f(\mathbf{h}\mathbf{W}_{\mathrm{down}})\mathbf{W}_{\mathrm{up}} = \mathbf{h} + \Phi(\mathbf{h}).
    \label{eq:adapter}
\]
% \vspace{-2mm}
\noindent \textbf{Target-only Adapter Tuning (Adapter FT)} trains a near-zero initialized target adapter $\Phi_t$ on $\mathcal{D}_t$ while keeping the Whisper backbone frozen. Two existing cross-lingual transfer strategies extend this baseline using a pre-trained source adapter. \textbf{Warm-initialized Transfer (Warm-init.)}, adapted from LoRA-Whisper~\cite{song2024lora}, initializes $\Phi_t$ from a source adapter $\Phi_s$ before training on $\mathcal{D}_t$. \textbf{Target-inclusive Fusion (Target-incl.)}, inspired by SimAdapter~\cite{hou2021exploit}, fuses pre-trained source and target adapters through layer-wise learned attention using a lightweight AdapterFusion variant~\cite{pfeiffer2021adapterfusion}. Our variant projects queries and keys to $d_k \ll d_{\mathrm{model}}$ and directly reweights adapter outputs without a value projection, keeping the additional parameter count comparable to a single adapter.

% ----------------------------------------
\subsection{Sequential Adapter Stacking}
\label{ssec:seqstack}
\begin{figure}[t]
\centering
\begin{subfigure}{.49\linewidth}
  \centering
  \includegraphics[width=.7\linewidth]{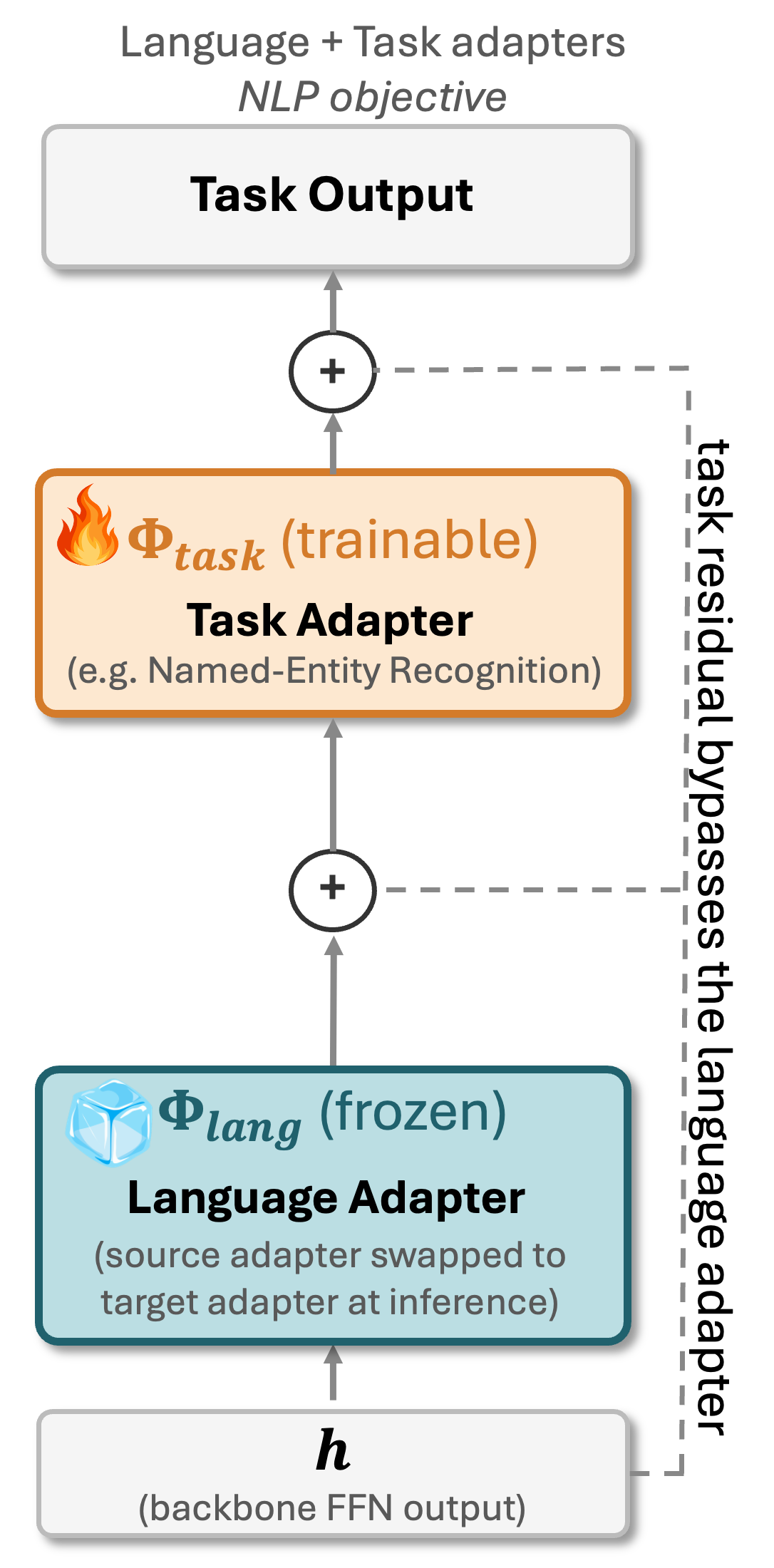}
  \caption{MAD-X Framework \cite{pfeiffer2020mad}}
  \label{fig:mad-x}
\end{subfigure}%
\begin{subfigure}{.49\linewidth}
  \centering
  \includegraphics[width=.7\linewidth]{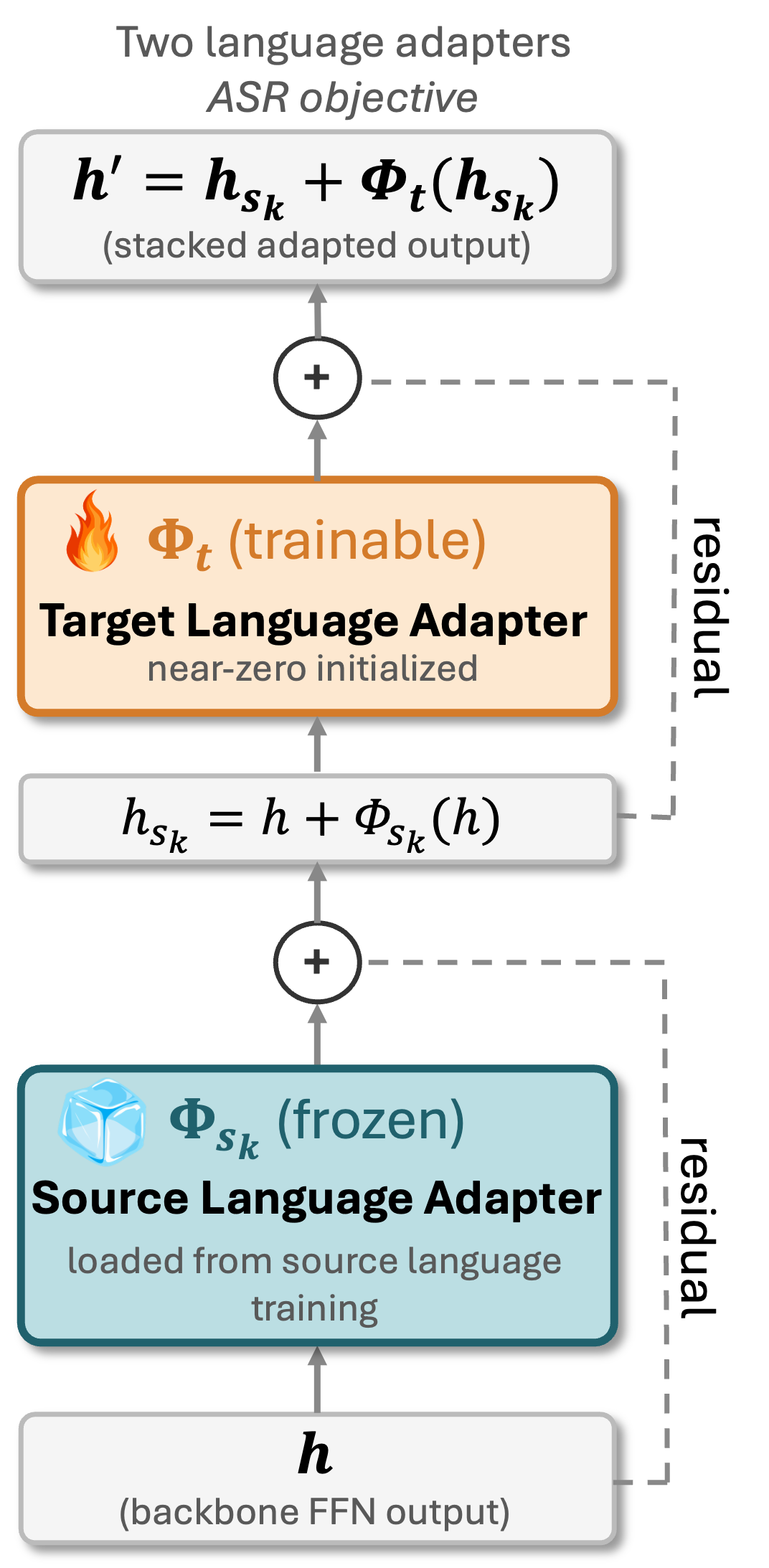}
  \caption{Sequential Stacking}
  \label{fig:sub2}
\end{subfigure}
\vspace{-3mm}
\caption{Framework of (a) MAD-X and (b) our proposed Sequential Stacking. Layer normalization is omitted for clarity.}
\label{fig:madx-seqstack}
\vspace{-5mm}
\end{figure}

% ----------------------------------------
% SeqStack vs. MAD-X
Motivated by the stacked-adapter design of MAD-X \cite{pfeiffer2020mad}, SeqStack composes source and target knowledge serially in the cross-lingual ASR setting. While MAD-X stacks a trainable \emph{task} adapter on a frozen language adapter, with the source language adapter replaced by the target language adapter at inference for zero-shot transfer, our approach stacks two \emph{language} adapters (Figure~\ref{fig:madx-seqstack}). A source language adapter $\Phi_{s_k}$ is first trained monolingually on $\mathcal{D}_{s_k}$ and frozen. A near-zero initialized target language adapter $\Phi_t$ is then stacked on top as the sole component optimized on $\mathcal{D}_t$:
% \vspace{-0.5cm}
\[
\mathbf{h}_{s_k}^{(l)} = \mathbf{h}^{(l)} + \Phi_{s_k}\big(\mathbf{h}^{(l)}\big),
\qquad
\mathbf{h}'^{(l)} = \mathbf{h}_{s_k}^{(l)} + \Phi_t\big(\mathbf{h}_{s_k}^{(l)}\big).
\]
% \vspace{-0.5cm}
Gradients propagate only through $\Phi_t$. The frozen source adapter provides a fixed, source-conditioned transformation of the backbone representation. The residual connection wiring also reflects each design's goal: MAD-X keeps its language adapters substitutable at inference, whereas SeqStack's source conditioning is meant to persist, so the target language adapter's residual wraps the source-conditioned state $\mathbf{h}_{s_k}$.

% ----------------------------------------

% ----------------------------------------

% ----------------------------------------
\subsection{Source Language Selection}
\label{ssec:langsec}
% ----------------------------------------

For each target, candidate sources are ranked before adaptation using three complementary measures of source--target relatedness computed from FLEURS \cite{conneau2023fleurs} text and speech.

% \noindent 
\textbf{Encoder representation similarity (EncSim).} This metric measures the similarity of Whisper encoder representations across languages using 100 utterances per language. First, hidden states for each utterance $i$ of language $\ell$ are mean-pooled over non-padding frames to extract layer-wise vectors, which are L2-normalized before averaging within four layer bands $\mathcal{B}$ (early 0--5, mid-shallow 6--11, mid-deep 12--17, and deep 18--23). To mitigate transformer anisotropy \cite{godey2024anisotropy}, the global mean across candidate languages is then subtracted before computing each language centroid $\mathbf{c}_\ell^{\mathcal{B}}$. Finally, the pairwise similarity is computed as the cosine distance between centroids: $\mathrm{sim}^{\mathcal{B}}(\ell_1, \ell_2) = (\mathbf{c}_{\ell_1}^{\mathcal{B}} \cdot \mathbf{c}_{\ell_2}^{\mathcal{B}}) / (\lVert \mathbf{c}_{\ell_1}^{\mathcal{B}} \rVert \lVert \mathbf{c}_{\ell_2}^{\mathcal{B}} \rVert)$.

% ----------------------------------------
% \begin{figure}[h]
%     \centering
%     \includegraphics[width=\columnwidth]{figs/encoder_heatmap.png}
%     \caption{Whisper encoder representation similarity by band, measured by centroid cosine similarity between language pairs.}
%     \label{fig:encoder-sim}
% \end{figure}
% ----------------------------------------

% \noindent 
\textbf{Token coverage (TokCov).} 
Decoder-side similarity is measured by Byte-Pair Encoding (BPE) token coverage, defined as the fraction of the target token stream appearing in the source corpus. Since Whisper's tokenization granularity is heavily influenced by pre-training exposure \cite{radford2022robust, liang2025beyond}, aggregate token coverage alone may conflate genuine lexical sharing with byte-level overlap. To mitigate this, tokens are classified by the number of complete Unicode code points they span (e.g. subword, character, or byte), and coverage is decomposed by script class (e.g. Latin, Brahmic, or CJK).

% \noindent 
\textbf{Genealogical relatedness (GenSim).} This model-\hspace{0pt}independent metric is measured by the cosine similarity between binary Glottolog family-tree vectors from URIEL~\cite{littell2017uriel}.

% \noindent 
All three measures produce the same source ranking for each target (Table~\ref{tab:selection}). Our experiments then compare the highest-ranked (close) source, the second-ranked (medium) source, and Mandarin as a distant control.

% ----------------------------------------
\begin{table}[t]
\centering
\caption{Target languages with candidate sources ranked by the similarity framework and Whisper pre-trained hours.}
\vspace{-3mm}
\label{tab:selection}
\begin{threeparttable}
% \resizebox{\columnwidth}{!}{%
\begin{tabular}{@{}l@{ }|@{ }l@{ }|@{ }r@{\hspace{4pt}}c@{\hspace{4pt}}c@{\hspace{4pt}}c@{}}
% \begin{tabular}{@{}l@{ }|@{ }l@{ }|@{ }rp{8.5mm}p{8.5mm}p{8.5mm}}
\toprule
% \textbf{Target} & \textbf{Source} & \textbf{Hours} &
% \makecell{\textbf{Token}\\\textbf{cov.\ (\%)}} &
% \makecell{\textbf{Enc.}\\\textbf{sim.}} &
% \makecell{\textbf{Gen.}\\\textbf{sim.}} \\
Target & Source & Hours & EncSim & TokCov(\%) & GenSim \\ 
\midrule
\multirow{3}{*}{\makecell[l]{Asturian}}
  & Spanish $\mathrm{es}$  & 11.1k & 0.81 & 80.2 & 0.96\\
  & French $\mathrm{fr}$  &  9.8k & 0.61 & 62.6 & 0.78 \\
  & Mandarin $\mathrm{zh}$ & 23.4k & 0.40 & --   & 0.00 \\
\midrule
\multirow{3}{*}{\makecell[l]{Assamese}}
  & Bengali $\mathrm{bn}$ & 1.3 & 0.78 & 27.2 & 0.80 \\
  & Hindi $\mathrm{hi}$   & 12.0  & 0.26 & 0.0  & 0.40 \\
  & Mandarin $\mathrm{zh}$ & 23.4k & -0.12 & 0.0 & 0.00 \\
\midrule
\multirow{3}{*}{\makecell[l]{Xhosa}}
  & Zulu  $\mathrm{zu}$   & 0.0   & 0.86 & 84.9 & 1.00 \\
  & Swahili $\mathrm{sw}$ & 5.4 & 0.71 & 76.1 & 0.61 \\
  & Mandarin $\mathrm{zh}$ & 23.4k & -0.38 & 27.5 & 0.00 \\
\bottomrule
\end{tabular}%
% }
\begin{tablenotes}
\footnotesize
\item EncSim: Deep-band centroid cosine similarity, where a value close to 1 indicates high similarity and -1 indicates maximal dissimilarity.
\end{tablenotes}
\end{threeparttable}
\vspace{-5mm}
\end{table}
% ----------------------------------------

% ----------------------------------------
\section{EXPERIMENTAL SETUP}
\label{sec:experiment}
% ----------------------------------------

% ----------------------------------------
% \subsection{Experimental Setup}
% \label{ssec:setup}
% ----------------------------------------

\textbf{Data and Languages.} The experiments are conducted on FLEURS \cite{conneau2023fleurs} across Asturian, Assamese, and Xhosa (Table~\ref{tab:target-dataset-splits}). These languages were selected because they share weak vanilla baselines (WER $\geq$ 50\%) while representing three distinct tiers of unsupported languages: those with extensive ($>$1,000 hours), partial (10--1,000 hours), and minimal ($<$10 hours) pre-training exposure of their closest related languages (Table~\ref{tab:selection}). For languages lacking a native Whisper language token, we use empirically related-language proxies: Spanish $\langle \mathrm{|es|} \rangle$ for Asturian and Swahili $\langle \mathrm{|sw|} \rangle$ for Xhosa. Although excluded from Whisper's ASR pre-training, Assamese utilizes its native token $\langle \mathrm{|as|} \rangle$ derived from the model's speech translation data. To control for training volume, source language adapters are each trained on approximately 120 hours of speech by combining each language's full FLEURS train split with data sampled from the train split of a supplementary corpus based on public availability. We use Common Voice \cite{ardila2020common} for Spanish, French, Mandarin and Swahili, IndicVoices \cite{javed2024indicvoices} for Bengali and Hindi, and Swivuriso \cite{marivate2025swivuriso} for Zulu.

% ----------------------------------------
\begin{table}[t]
\centering
\caption{FLEURS dataset statistics for target languages.}
\vspace{-3mm}
\label{tab:target-dataset-splits}
\begin{tabular}{llrrr}
\toprule
Language & Lang. Token & Train (h) & Val (h) & Test (h) \\
\midrule
Asturian & $\langle \mathrm{|es|} \rangle$ & 7.5  & 0.9 & 2.4 \\
Assamese & $\langle \mathrm{|as|} \rangle$ & 10.7 & 1.4 & 3.5 \\
Xhosa    & $\langle \mathrm{|sw|} \rangle$ & 13.3 & 1.5 & 3.8 \\
\bottomrule
\end{tabular}
\vspace{-5mm}
\end{table}
% ----------------------------------------

% \noindent 
\textbf{Pre-processing.} Audio is resampled to 16kHz and filtered to 0.5--30 seconds. Text normalization follows a two-tier design applied to both references and hypotheses before scoring, covering standard normalization (Unicode NFC, lowercasing, punctuation removal) and language-specific orthographic rules to preserve diacritics and elision.

% \noindent 
\textbf{Training and Decoding Configurations.} Whisper medium is utilized to balance computational efficiency and performance. To control for the trainable parameter count across adapter-based methods, Pfeiffer adapters adopt dimension $d=256$, and fusion layers use projection dimension $d_k=256$. Models are optimized using AdamW with BF16 precision, an effective batch size of 32, and linear learning-rate decay with 5\% warmup. Based on a small grid search, peak learning rates are set to $1\times 10^{-5}$ for full fine-tuning, $3\times 10^{-4}$ for adapter methods, and $5\times 10^{-5}$ for fusion. Training runs up to 20 epochs with an early-stopping patience of 3 epochs based on validation WER. Decoding uses greedy search without external language model re-scoring. 

% \noindent 
\textbf{Evaluation Metric.} We report Word Error Rate (WER) and determine statistical significance using the Matched-Pairs Sentence-Segment Word Error test (MAPSSWE) at $p<0.05$.

% ----------------------------------------
\section{Results and Discussion}
\label{sec:results}
% \subsection{Low-resource Language Adaptation Results}
% \label{ssec:results}
% ----------------------------------------

\textbf{Baseline Performance.}
Vanilla Whisper achieves reasonable performance on Asturian with a WER of 52.5\% but collapses on Xhosa and Assamese ($>$130\% WER) (Table~\ref{tab:low-resource-compare}). Although monolingual adapter tuning substantially improves performance, full fine-tuning establishes a stronger reference for evaluating cross-lingual transfer methods, reducing WER from the vanilla baseline by 70\% on Asturian (15.7\%), 74\% on Assamese (36.3\%), and 69\% on Xhosa (41.9\%).

% \noindent 
\textbf{Source-Adapter Transfer Evaluation.} Across all source--target pairs, \emph{Warm-initialized Transfer} performs inconsistently against adapter tuning, improving on some pairs while degrading on others. The inconsistency persists even with the closest source, Asturian--Spanish, which falls short of full fine-tuning. This likely occurs because warm initialization primarily provides a better optimization starting point. Since monolingual adapter tuning already achieves strong performance from a near-identity initialization, the head start from source-learned weights provides negligible benefit. 

\emph{Target-inclusive Fusion} outperforms adapter tuning on certain pairs with a close source (Assamese--Bengali and Xhosa--Zulu) but lags behind full fine-tuning for all targets. On Asturian--Spanish, it degrades performance below adapter tuning. To understand this regression, we analyze how the fusion mechanism allocates attention weights across the frozen adapters. In the deep band, decoder attention correctly exceeds 70\% on the Asturian adapter, whereas encoder attention drifts down to 40\%. %exhibits a problematic drift, dropping attention to the Asturian adapter to 40\% in the deep band. 
Since adapter tuning is already close to full fine-tuning, this persistent reliance on the Spanish adapter dilutes a near-ceiling system rather than adding useful knowledge. Prior works address similar attention drift using fusion guide loss \cite{hou2021exploit} or dot-product mixture weighting \cite{hucam2024crosslingual}. However, these methods were validated on backbones with limited or no multilingual pre-training, leaving their effectiveness in our setting open. Under a constrained compute budget, we prioritize a structural alternative (SeqStack) that sidesteps the allocation problem rather than regularizing it.

% ----------------------------------------
\begin{table}[t]
\centering
% \caption{FLEURS test WER (\%) of low-resource language adaptation.}
\caption{Low-resource language adaptation on FLEURS (\%WER).}
\vspace{-3mm}
\label{tab:low-resource-compare}
\begin{threeparttable}
\begin{tabular}{@{}l@{ }|l|ccc@{}}
\toprule
\textbf{Method} & \textbf{Source}
& \textbf{Asturian} & \textbf{Assamese} & \textbf{Xhosa} \\
\midrule
Vanilla & -- & 52.5 & 141.2 & 133.3 \\
Full FT & -- & 15.7 & 36.3 & 41.9 \\
Adapter FT   & -- & 16.8 & 40.5 & 49.6 \\
\midrule
\multirow{3}{*}{Warm-init.}
 & Close  & 16.6 & \textbf{33.6}† & \textbf{38.5}† \\
 & Medium  & 17.3 & 39.1 & 45.4 \\
 & Distant & 16.8 & 42.3 & 44.1 \\
 \cmidrule{1-5}
\multirow{3}{*}{Target-incl.}
 & Close    & 20.4 & 39.7 & 45.1 \\
 & Medium   & 16.9 & 42.6 & 55.0 \\
 & Distant  & 16.9 & 40.2 & 51.0 \\
 \cmidrule{1-5}
\multirow{3}{*}{SeqStack}
 & Close   & \textbf{14.9}† & 34.1† & \textbf{38.5}† \\
 & Medium  & 15.9 & 38.1 & 44.3 \\
 & Distant & 16.3 & 40.4 & 44.3 \\
\bottomrule
\end{tabular}
\end{threeparttable}
\begin{tablenotes}
\footnotesize
\item Close/medium/distant sources are $\mathrm{es}/\mathrm{fr}/\mathrm{zh}$ for Asturian, $\mathrm{bn}/\mathrm{hi}/\mathrm{zh}$ for Assamese, and $\mathrm{zu}/\mathrm{sw}/\mathrm{zh}$ for Xhosa.
\item † denotes $p<0.05$ vs. Full fine-tuning (Full FT).
\end{tablenotes}
\vspace{-5mm}
\end{table}
% ----------------------------------------

% \noindent 
\textbf{Sequential Adapter Stacking Evaluation.} Across all targets, SeqStack with the target's closest source language -- Spanish for Asturian, Bengali for Assamese, and Zulu for Xhosa -- improves significantly over both adapter tuning and full fine-tuning. The relative margin over full fine-tuning grows as related-language pre-training coverage shrinks (5\% on Asturian, 6\% on Assamese, and 8\% on Xhosa). Furthermore, SeqStack's performance validates our three-axis similarity framework. The framework identifies the relative order of candidate sources, and empirical results demonstrate that the closest source consistently performs better than both medium and distant controls. On Xhosa, where the medium and distant sources tie, we conduct an ablation to understand whether the performance gains stem from cross-lingual transfer or the raw parameter increase of stacking adapters. Doubling the capacity of the baseline monolingual adapter ($d=512$) reduces WER from 49.6\% to 45.6\%, whereas SeqStack with the Zulu adapter at our controlled dimension $d=256$ yields 38.5\% WER. This suggests that language-specific transfer drives the improvement on Xhosa--Zulu.

% However, the similarity framework does not predict transfer magnitude. 
While the closest source consistently outperforms less related ones under SeqStack, medium sources do not reliably improve upon the distant control. We hypothesize that this stems from two factors. French's Romance-general knowledge is plausibly redundant with Whisper's extensive Romance pre-training, leaving Spanish's advantage for Asturian branch-specific. Conversely, Xhosa's lack of transfer from Swahili highlights a genealogical boundary, favoring Zulu, a closely related Nguni language, over a distant Bantu branch. Therefore, the similarity framework serves best as an ordinal instrument for closest source selection rather than a predictor of transfer magnitude. Overall, warm-initialized and fusion-based methods perform inconsistently across source--target language pairs, whereas SeqStack empirically does not fall below adapter tuning performance on any pairs. When the best source language is uncertain, SeqStack is a safer strategy for extending multilingual ASR to unsupported languages.

\begin{table}[t]
\centering
\caption{FLEURS test WER (\%) on one-hour target-data.}
\vspace{-3.5mm}
\label{tab:ablation}
\begin{threeparttable}
\begin{tabular}{lccc}
\toprule
\textbf{Method} & \textbf{Asturian} & \textbf{Assamese} & \textbf{Xhosa} \\
% Method & Asturian & Assamese & Xhosa \\
\midrule
Full FT & \textbf{21.3} & 53.6 & 63.9 \\
Adapter FT & 23.1 & 100.0 & 70.0 \\
SeqStack (Close) & 21.8 & \textbf{47.0}† & \textbf{47.5}† \\
\bottomrule
\end{tabular}
\end{threeparttable}
\begin{tablenotes}
    \footnotesize
    \item † denotes $p<0.05$ vs. Full FT.
\end{tablenotes}
\vspace{-5mm}
\end{table}
% ----------------------------------------
\textbf{Low-regime Analysis.}
To evaluate data efficiency under extreme low-resource conditions, we restrict target training data to a one-hour subset and compare SeqStack, paired with each target's closest source, against full fine-tuning and adapter tuning trained on the same subset. The source adapters remain trained on their full 120-hour corpora. Full fine-tuning is functional across all targets. Adapter tuning under-performs substantially and collapses completely on Assamese (Table~\ref{tab:ablation}). %, generating repetitive byte loops (Table~\ref{tab:ablation}). % This is likely due to Assamese fragments into byte-level tokens (Section~\ref{ssec:langsec}), yielding target sequences $6\times$ longer than Latin-script equivalents. 
In contrast, SeqStack largely maintains its advantage, achieving relative WER reductions over full fine-tuning of 12\% on Assamese and 26\% on Xhosa. The frozen source adapter provides both cross-lingual transfer and the architectural stability that adapter tuning lacks.

% While full fine-tuning remains functional across all targets, monolingual adapter tuning falls behind and collapses completely on Assamese into repetitive byte loops. This failure likely stems from tokenization, where Assamese fragments into byte-level tokens (Section~\ref{ssec:langsec}), creating target sequences $6\times$ longer than Latin-script equivalents. A small monolingual adapter given only one hour of data struggles to learn these long chains from scratch. In contrast, SeqStack circumvents this by inheriting a frozen Bengali module already capable of generating well-formed Eastern Nagari sequences. Ultimately, empirical results under extreme low-resource conditions indicate that the frozen source adapter provides both cross-lingual transfer and, as seen on Assamese, the architectural stability that adapter tuning lacks.
\vspace{-0.5mm}

% ----------------------------------------
\section{CONCLUSION}
\label{sec:conclusion}
% ----------------------------------------

This work investigated adapting massively multilingual ASR backbones to unseen languages by optimizing both source selection and transfer mechanisms. Under controlled experiments, the proposed SeqStack proves robust across source--target pairs, mitigating the inconsistencies observed on warm-initialized and fusion-based transfer methods. Paired with each target's closest source, SeqStack significantly outperforms full fine-tuning, reducing WER by 5--8\% on full training data and 12--26\% on a one-hour subset. The gain is most substantial on Xhosa, where the backbone's related-language coverage is weakest, and language-specific transfer materializes only with the closest related source. %While SeqStack is architecturally applicable to other adapter-compatible speech foundation models, our experiments are limited to Whisper due to the computational scope of this study, evaluating its generality across different pretrained backbones remains future work.
While SeqStack is architecturally applicable to other adapter-compatible speech foundation models, we focus on Whisper due to computational constraints, leaving its generalizability to other backbones for future work.

% We demonstrate that the proposed SeqStack consistently outperforms full fine-tuning, yielding relative WER reductions of 5--8\% on full data and 12--26\% with one hour of labeled target data. Cross-lingual transfer benefits are largest where the backbone's related-language coverage is weakest, provided the source belongs to the same family branch and ranks highly on our three-axis similarity framework.

% This work investigated whether parameter-efficient adapter composition can mitigate performance degradation of multilingual ASR backbones on unseen languages. Our empirical results demonstrate that SeqStack offers the most benefit precisely where the backbone's support is weakest and that language-specific transfer materializes only with a closely related source from the same language branch. SeqStack yields 5--8\% WER reduction under full data-regime and 12--26\% relative improvement with only 1 hour of labeled data over full fine-tuning. The backbone-support taxonomy combined with our language similarity framework also provides a practical heuristic for source language selection when extending multilingual ASR systems to unsupported languages.

% ----------------------------------------

% ----------------------------------------
% \vfill\pagebreak
\clearpage

% \section{REFERENCES}
% \label{sec:refs}

% Please follow the IEEE Citation Guidelines, \url{https://ieee-dataport.org/sites/default/files/analysis/27/IEEE\%20Citation\%20Guidelines.pdf} for formatting of references.

\bibliographystyle{IEEEbib}
\bibliography{strings,refs}

\end{document}